\def\year{2021}\relax
\documentclass[letterpaper]{article} % DO NOT CHANGE THIS
\usepackage{aaai21}  % DO NOT CHANGE THIS
\usepackage{times}  % DO NOT CHANGE THIS
\usepackage{helvet} % DO NOT CHANGE THIS
\usepackage{courier}  % DO NOT CHANGE THIS
\usepackage[hyphens]{url}  % DO NOT CHANGE THIS
\usepackage{graphicx} % DO NOT CHANGE THIS
\usepackage{natbib}  % DO NOT CHANGE THIS AND DO NOT ADD ANY OPTIONS TO IT
\usepackage{caption} % DO NOT CHANGE THIS AND DO NOT ADD ANY OPTIONS TO IT
\usepackage{subfig}
\usepackage{amsmath}    
\usepackage{amsfonts}    
\usepackage{booktabs}
\usepackage{multirow}
\usepackage{enumitem}
\usepackage{colortbl}
\usepackage{import}

\newcommand{\eg}{\textit{e.g.}}
\newcommand{\norm}[1]{\left\lVert#1\right\rVert}

\makeatletter
\newcommand{\printfnsymbol}[1]{%
  \textsuperscript{\@fnsymbol{#1}}%
}

\usepackage{kotex}

\makeatother

\title{Multi-level Distance Regularization for Deep Metric Learning}
\author{
    \hspace{-1cm}
    Yonghyun Kim\textsuperscript{\rm 1}\thanks{Equal Contribution} \hspace{3cm} Wonpyo Park\textsuperscript{\rm 2}\printfnsymbol{1}
    \\
}
\affiliations{
    \hspace{-2.0cm}
    \textsuperscript{\rm 1}AI Lab, Kakao Enterprise \hspace{3.3cm} %\qquad
    \textsuperscript{\rm 2}Kakao Corp. \\
    \hspace{-1.2cm} aiden.d@kakaoenterprise.com \hspace{2.3cm} tony.nn@kakaocorp.com

}
\begin{document}

\maketitle

\begin{abstract}
We propose a novel distance-based regularization method for deep metric learning called Multi-level Distance Regularization (MDR). 
MDR explicitly disturbs a learning procedure by regularizing pairwise distances between embedding vectors into multiple levels that represents a degree of similarity between a pair. 
In the training stage, the model is trained with both MDR and an existing loss function of deep metric learning, simultaneously; the two losses interfere with the objective of each other, and it makes the learning process difficult. 
Moreover, MDR prevents some examples from being ignored or overly influenced in the learning process.
These allow the parameters of the embedding network to be settle on a local optima with better generalization.
Without bells and whistles, MDR with simple Triplet loss achieves the-state-of-the-art performance in various benchmark datasets: CUB-200-2011, Cars-196, Stanford Online Products, and In-Shop Clothes Retrieval.
We extensively perform ablation studies on its behaviors to show the effectiveness of MDR.
By easily adopting our MDR, the previous approaches can be improved in performance and generalization ability.
\end{abstract}

\section{Introduction}

% DML에 대한 Introduction
Deep Metric Learning (DML) aims to learn an appropriate metric that measures the semantic difference between a pair of images as a distance between embedding vectors.
Many research areas such as image retrieval \cite{sohn2016improved,yuan2017hard,oh2017deep,duan2018deep,ge2018deep} and face recognition \cite{NormFace,SphereFace,CosFace,ArcFace} are based on DML to seek appropriate metrics among instances.
Those studies focus on devising a better loss function for DML.

%removed suh2019stochastic

\begin{figure*}[t]
	\begin{center}
		\includegraphics[height=5.2cm]{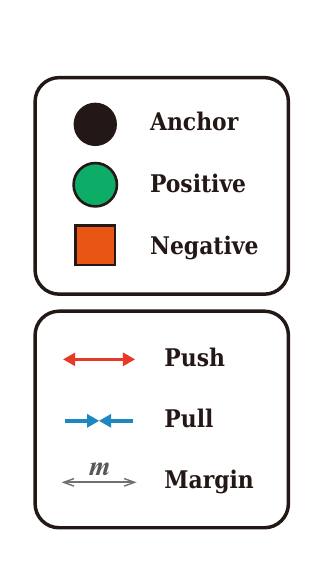}
		\subfloat[Conventional Learning]{
			\label{fig:concept_a}
			\includegraphics[height=5.2cm]{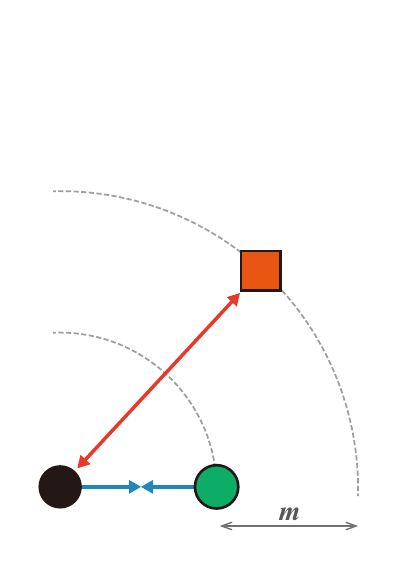}
		}
		\subfloat[Our Learning]{
			\label{fig:concept_b}
			\includegraphics[height=5.2cm]{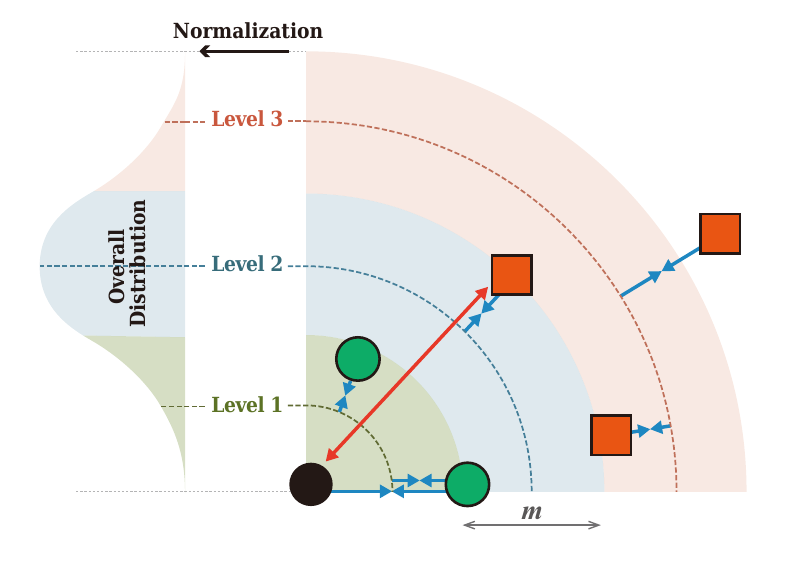}
		}	
	\end{center}
	\caption{ 
	Conceptual comparison between the conventional learning scheme and our learning scheme.
  (a) illustrates the triplet learning \cite{schroff2015facenet}, which is one of the representative conventional learning.
  It increases the relative difference between distances of a positive pair and that of a negative pair more than margin $m$.
(b) illustrates our learning combined with the triplet learning.
It has multiple levels with disjoint intervals to reflect various degrees of similarity between pairs.
It disturbs the learning procedure to construct an efficient embedding space by preventing the pairwise distances from deviating from its belonging level.
}
	\label{fig:concept}
\end{figure*}

% Brief Previous Works and Limitations
Most of previous loss functions \cite{sohn2016improved,bromley1994signature,hadsell2006dimensionality,yideep2014,hoffer2015deep,schroff2015facenet} use binary supervision that indicates whether a given pair is positive or negative.
Their common objective is to minimize the distance between a positive pair and maximize the distance between a negative pair (Figure \ref{fig:concept_a}).
However, without any constraints, a model trained with such objective is prone to overfitting on a training set because positive pairs can be aligned too closely while the negative pairs can be aligned too far in the embedding space.
Therefore, several loss functions employ additional terms to avoid positive pairs to be too close and negative pairs to be too far, \eg, margin $m$ in Triplet loss \cite{schroff2015facenet} and Constrastive loss \cite{hadsell2006dimensionality}.
Despite these attempts, they still can suffer from overfitting due to the lack of explicit regularization for the distances.

Our insight is that a learning procedure of DML can be enhanced by explicitly regularizing the distance between pairs to disturb a loss function of DML from optimizing an embedding network;
one easy way to constrain a distance is to pull the value of the distance to a predefined level.
Conventional loss functions of DML adjust the distance according to its label, on the other hand, explicit distance-based regularization prevents the distance from deviating from the predefined level.
Those two interfere with the objective of each other, thus it makes the learning process difficult and allows the embedding network to be more robust for generalization.
Additionally, we consider multiple levels with disjoint intervals to regularize distances, not a single level, because a degree of inter-class similarity or intra-class variation can be different depending on classes or instances.

% Proposed Method
We propose a novel method called Multi-level Distance Regularization (MDR) that makes the conventional loss functions of DML have difficulty in converging by holding each distance so that it does not deviate from the belonging level.
At first, MDR normalizes pairwise distances among the embedding vectors of a mini-batch, with their mean and standard deviation to obtain the objective degree of similarity between a pair by considering overall distribution.  
MDR defines the multiple levels that represent various degrees of similarity for pairwise distances, and the levels and the belonging distances are trained to approach each other (Figure \ref{fig:concept_b}).
A conventional loss function of DML struggles to optimize a model by overcoming the disturbance from the proposed regularization. 
Therefore, the learning process succeeds in learning a model with a better generalization ability.
We summarize our contributions:

\begin{itemize}[leftmargin=*]
    \item  
    We introduce MDR, a novel regularization method for DML. The method disturbs optimizing pairwise distances by preventing them from deviating from its belonging level for better generalization.
    \item 
    MDR achieves the-state-of-the-art performance on various benchmark datasets \cite{CUB-200,Cars-196,Song2016DeepML,InShop} of DML.
    Moreover, our extensive ablation studies show that MDR can be adopted to any backbone networks and any distance-based loss functions to improve the performance of a model.
    % of a given model
\end{itemize}

\section{Related Work}

\noindent\textbf{Loss Function.}
Improving the loss function is one of the key objectives in recent DML studies.
One family of loss functions \cite{sohn2016improved,bromley1994signature,schroff2015facenet,Song2016DeepML,wang2019multi,wu2017sampling} focuses on optimizing pairwise distance between instances.
The common objective of these functions is to minimize the distance between positive pairs and to maximize the distance between negative pairs in an embedding space.
Contrastive loss \cite{bromley1994signature} samples pairs of two instances, whereas Triplet loss \cite{schroff2015facenet} samples triplets of anchor, positive and negative instances; then both losses optimize the distance between the sampled instances.
%% Added Start
Also, Global Loss \cite{kumar2016learning} minimizes the mean and variance of all pairwise distances between positive examples and maximizes the mean of pairwise distances between all negative examples; Global Loss helps to optimize examples that are not selected by the example mining of DML. 
Histogram Loss \cite{ustinova2016learning} minimizes the probability that a randomly sampled positive pair has a smaller similarity than randomly sampled negative pairs. 
%% Added End
To extend the number of relations explored at once, NPair \cite{sohn2016improved} samples a positive and all negative instances for each example in a given mini-batch; similar loss functions \cite{Song2016DeepML,wang2019multi} also sample a large number of instances to fully explore the pairwise relations in the mini-batch. 
On the other, some loss functions \cite{cakir2019deep,revaud2019learning} focus on learning to rank according to the similarity between pairs. 
The performance of loss functions optimizing pairwise distance can be changed by a sampling method, thus, several studies focused on the pair sampling \cite{suh2019stochastic,schroff2015facenet,wu2017sampling} for stable learning and better accuracy.
A recent work \cite{wang2020cross} even samples pairs across mini-batches to collect a sufficient number of negative examples.
Instead of designing a sampling method manually, a work \cite{roth2020pads} employs reinforcement learning to learn the policy for sampling.
As a regularizer, MDR can be combined with those loss functions to improve the generalization ability of a model.
\smallbreak

%Global Loss minimizes the mean and variance of all pairwise distances between positive examples and maximizes the mean of pairwise distances between all negative examples; Global Loss helps to optimize examples that are not selected by the example mining of DML. Histogram Loss minimizes the probability that a randomly sampled positive pair has a smaller similarity than randomly sampled negative pairs. Those two methods use the distribution of the pairwise similarities similar to MDR. However, they are not devised as a regularizer, rather they aim to optimize the overall distribution of pairwise similarity between all examples in a mini-batch, like conventional DML losses. Therefore, Global Loss and Histogram Loss can be combined with MDR to improve the performance of a model. We will clarify the difference in the related works.

\begin{figure*}
  \centering
  \includegraphics[width=15cm]{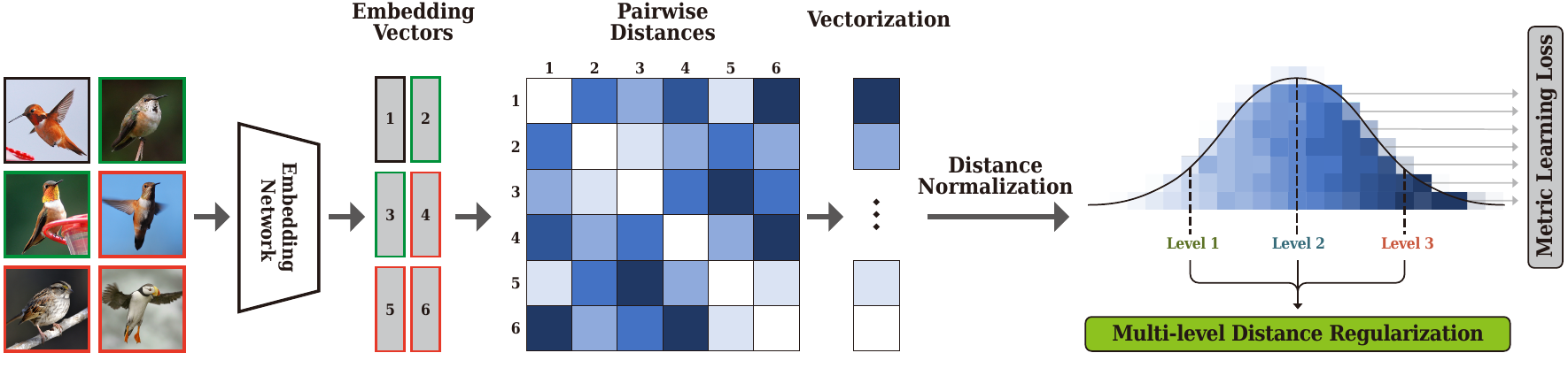}
  \caption{
    Learning procedure of the proposed MDR.
    The embedding network generates embedding vectors from given images.
    Our MDR computes a matrix of pairwise distances for the embedding vectors, and then, the distances are normalized after vectorization.
  In our learning scheme, a model is trained by simultaneously optimizing the conventional metric learning loss such as Triplet loss \cite{schroff2015facenet} and the proposed loss, which regularizes the normalized pairwise distances with multiple levels.
    }
	\label{fig:scheme}
\end{figure*}

\noindent\textbf{Generalization Ability.} Another goal of DML is to improve the generalization ability of a given model.
An ensemble of multiple heads that share the backbone network \cite{opitz_2018_pami,Kim_2018_ECCV,JACOB_2019_ICCV,sanakoyeu2019divide} has the key objective of diversifying each head to achieve reliable embedding.
Boosting can be used to re-weight the importance of instances differently on each head \cite{opitz_2018_pami,sanakoyeu2019divide},
or a spatial attention module can be used to differentiate a spatial region on which each head focuses \cite{Kim_2018_ECCV}.
HORDE \cite{JACOB_2019_ICCV} makes each head approximate a different higher-order moment.
Those methods focus on changing the architecture of a model, but our MDR, as a regularizer, focuses on making a learning procedure harder to improve generalization ability. 
Without adding any extra computational costs or changing the architecture of the model, MDR can be easily integrated with those DML methods by simply adding our loss function.

\section{Proposed Method}

In this section, we introduce a new regularization method called Multi-level Distance Regularization (MDR), which makes the learning procedure difficult by preventing each pairwise distance from deviating from a corresponding level, to learn a robust feature representation.

\subsection{Multi-level Distance Regularization}

We describe the detailed procedure of MDR to regulate pairwise distances in three steps (Figure \ref{fig:scheme}).
\smallbreak

\noindent\textbf{(1) Distance Normalization.} 
This step is performed to obtain an objective degree of distance by considering overall distribution for stable regularization.
Here, an embedding network $f$ maps an image $x$ into an embedding vector $e$ with a certain dimensionality: $e = f(x)$.
A distance is defined as Euclidean distance between two given embedding vectors, $d(e_i,e_j)=\norm{e_i-e_j}_2$.
We normalize the distance as: 
\begin{equation}
    \bar{d}(e_i, e_j) = \frac{d(e_i, e_j) - \mu}{\sigma},
\end{equation}
where $\mu$ is mean of distances and $\sigma$ is standard deviation of distances a set of pairs, which is $\mathcal{P} = \left\{(e_i, e_j) | i \neq j \right\}$ for all instances of a mini-batch.
To more widely consider the overall dataset, we employ the momentum updates:
\begin{equation}
\begin{aligned}
    \mu^{*}_t = \gamma \mu^{*}_{t-1} + (1 - \gamma) \mu, \\
    \sigma^{*}_t = \gamma \sigma^{*}_{t-1} + (1 - \gamma) \sigma,
\end{aligned}
\end{equation}
where $\mu^{*}_t$ and $\sigma^{*}_t$ are respectively the momented mean and momented standard deviation at iteration $t$, and $\gamma$ is the momentum.
With the momented statistics, the normalized distance is re-written:
\begin{equation}
    \bar{d}(e_i, e_j) = \frac{d(e_i, e_j) - \mu^{*}}{\sigma^{*}}.
\end{equation}
%% Added Start
%Our distance normalization is functionally similar to Batch Normalization \cite{ioffe2015batch} since it also normalizes the range of pairwise distances using mean and standard deviation. However, our normalization is not a layer inserted into a neural network like Batch Normalization, rather it is a pre-processing to prevent MDR loss from diverging.
\smallbreak
%% Added End

%In conventional learning, Batch Normalization helps network training by constraining the range of values to reduce internal covariate shift. Our distance normalization is functionally similar to Batch Normalization since it also normalizes the range of pairwise distances using mean and standard deviation. However, our normalization is not a layer inserted into a neural network like Batch Normalization, rather it is a pre-processing to prevent MDR loss from diverging. As mentioned in the reviews, our normalization can be replaced with BN, functionally. Both show almost the same results in CUB-200; MDR (69.1256), MDR-BN (68.8555).

\noindent\textbf{(2) Level Assignment.}
MDR designates a level that acts as a regularization goal for each normalized distance.
We define a set of levels $s\in\mathcal{S}$, and the levels are initialized with predefined values; 
each level $s$ is interpreted as a multiplier of the standard deviation of the normalized distance.
$g(d; s)$ is an assignment function that outputs whether the given distance $d$ and the given level $s$ are the closest or not, and is defined as:
\begin{equation}
g(d, s) =
\begin{cases}
  1, & \text{if } \arg \min _ {s_{i}\in\mathcal{S}} | d - s_i | \text{ is } s
\\
  0, & \text{otherwise}.
\end{cases}
\end{equation}
By adopting the assignment function, MDR selects valid regularization levels for each distance with the consideration of various degrees of similarities.
\smallbreak

\noindent\textbf{(3) Regularization.}
Finally, this step is performed to prevent pairwise distances from deviating from its belonging level.
MDR minimizes the difference between a given normalized pairwise distance and the assigned level:
\begin{equation}
\begin{aligned}
    \mathcal{L}_{\text{MDR}} = \frac{1}{\mathcal{P}}\sum_{(e_i, e_j) \in \mathcal{P}}\sum_{s\in\mathcal{S}} g\left(\bar{d}(e_i, e_j),s\right)
    \cdot\left|\bar{d}(e_i, e_j)-s\right|.
    %\\ + \nu \sum_{s \in \mathcal{S}} s^{2},
    \label{eq:reg}
\end{aligned}
\end{equation}
%where $\nu$ is a hyper-parameter that determines the effects of the second term.
The levels are learnable parameters and are updated to optimally regularize the pairwise distances.
%In the first term, each normalized distance is trained to become closer to the assigned level; the assigned level is also trained to become closer to the corresponding distances.
Each normalized distance is trained to become closer to the assigned level; the assigned level is also trained to become closer to the corresponding distances.
As iterations pass, the levels are trained to properly divide the normalized distances into multiple intervals.
Each level is a representative value of a certain interval in the normalized distance.
We describe the initial configuration of the levels in Section \ref{sec:abl}.
%The second term in Eq. \ref{eq:reg} regularizes $s$ to prevent divergence of the levels.
\smallbreak

In conclusion, MDR has two functional effects of regularization:
(1) the multiple levels of MDR disturbs optimizing the pairwise distances among examples,
(2) the outermost levels of MDR prevents the positive pairs from getting too close and the negative pairs from getting too far.
By the formal effect, the learning process does not easily suffer from overfitting.
By the latter effect, the learning process does not suffer from diminishing of the loss from easy examples, and also, does not suffer from being too biased to certain examples such as hard examples.
Therefore, MDR stabilizes the learning procedure to achieve a better generalization ability on a test dataset.

\subsection{Learning}

\begingroup
\newcolumntype{C}[1]{>{\centering\let\newline\\\arraybackslash\hspace{0pt}}m{#1}}
\begin{table*}[t]
	\begin{center}
		\subfloat[CUB-200 \cite{CUB-200} and Cars-196 \cite{Cars-196} \label{tab:exp_a} ]
		{
			\centering
			\small
			\def\arraystretch{1.2}
            %\begin{tabular}{p{3.6cm}|C{0.8cm}C{0.8cm}C{0.8cm}C{0.8cm}|C{0.8cm}C{0.8cm}C{0.8cm}C{0.8cm}}
            \begin{tabular}{l|cccc|cccc}
               \hline
                  &  \multicolumn{4}{c|}{CUB-200} & \multicolumn{4}{c}{Cars-196} \\
               \cline{2-5}\cline{6-9}
                  Recall@\textit{K} & 1 & 2 & 4 & 8 & 1 & 2 & 4 & 8  \\
                \hline
                %Contrastive \cite{hadsell2006dimensionality} &  - & - & - & - & - & - & - & - \\
                %Margin \cite{wu2017sampling} &  - & - & - & - & - & - & - & - \\
                %NPair \cite{wu2017sampling} &  - & - & - & - & - & - & - & - \\
             HTL \cite{ge2018deep} &  57.1 & 68.8 & 78.7 & 86.5 & 81.4 & 88.0 & 92.7 & 95.7 \\
                RLL-H \cite{wang2019ranked} &   57.4 & 69.7 & 79.2 & 86.9 & 74.0 & 83.6 & 90.1 & 94.1 \\
                 NSM~\cite{zhaiclassification2019} & 59.6 & 72.0 & 81.2 & 88.4 & 81.7 & 88.9 & 93.4 & 96.0 \\
                MS \cite{wang2019multi} &  65.7 & 77.0 & 86.3 & 91.2 & 84.1 & 90.4 & 94.0 & 96.5
             \\
                SoftTriple \cite{qian2019softtriple} &  65.4 & 76.4 & 84.5 & 90.4 & 84.5 & 90.7 & 94.5 & 96.9 \\
                %Proxy-NCA \cite{wu2017sampling} &  - & - & - & - & - & - & - & - \\
                HORDE$^\dagger$ \cite{JACOB_2019_ICCV} &  66.3 & 76.7 & 84.7 & 90.6 & 83.9 & 90.3 & 94.1 & 96.3 \\
                %Proxy-Anchors \cite{kim2020proxy} &  68.4 & 79.2 & 86.8 & 91.6 & 86.1 & 91.7 & 95.0 & 97.3 \\
                DiVA \cite{milbich2020diva} &  66.8 & 77.7 & - & - & 84.1 & 90.7 & - & - \\
                \rowcolor[gray]{.9}
                Triplet & $57.3_{\pm0.7}$	& $68.7_{\pm0.8}$ & $78.4_{\pm0.6}$	& $86.1_{\pm0.4}$
                & $76.2_{\pm0.6}$	& $84.4_{\pm0.3}$ & $90.0_{\pm0.2}$	& $93.7_{\pm0.2}$ \\
                \rowcolor[gray]{.9}    
                Triplet+$L_2$Norm &  
                $65.1_{\pm0.3}$ & $76.1_{\pm0.2}$ & $84.2_{\pm0.2}$ & $90.3_{\pm0.1}$ &
                $79.8_{\pm0.3}$ & $87.1_{\pm0.3}$ & $91.9_{\pm0.4}$ & $95.1_{\pm0.1}$\\
                \rowcolor[gray]{.9}
                %Ours$^{512}$ & 68.4 & 78.2 & 85.6 & 91.2 & 87.7 & 92.6 & 95.3 & 97.3 \\
                Triplet+MDR &
                $\textbf{68.8}_{\pm0.5}$ & $\textbf{78.8}_{\pm0.3}$ & $\textbf{86.6}_{\pm0.2}$ & $\textbf{91.8}_{\pm0.1}$ &
                $\textbf{88.5}_{\pm0.3}$ & $\textbf{93.0}_{\pm0.2}$ & $\textbf{95.6}_{\pm0.2}$ & $\textbf{97.5}_{\pm0.1}$
                \\
                \rowcolor[gray]{.9}
                Triplet+MDR$^\dagger$&  
                $\textbf{71.4}_{\pm0.4}$ &$\textbf{81.2}_{\pm0.3}$ & $\textbf{88.0}_{\pm0.2}$ & $\textbf{92.6}_{\pm0.3}$ &
                $\textbf{90.4}_{\pm0.2}$ & $\textbf{94.3}_{\pm0.1}$ & $\textbf{96.6}_{\pm0.1}$ &
                $\textbf{98.0}_{\pm0.1}$ \\  
                \hline
               \end{tabular}
		}		
		\\
		\subfloat[SOP \cite{Song2016DeepML} and In-Shop \cite{InShop}\label{tab:exp_b} ]
		{
			\centering
			\small
			\def\arraystretch{1.2}
        	\begin{tabular}{l|cccc|cccc}
           \hline
              &  \multicolumn{4}{c|}{SOP} & \multicolumn{4}{c}{In-Shop} \\
           \cline{2-5}\cline{6-9}
              Recall@\textit{K} & 1 & 10 & 100 & 1000 & 1 & 10 & 20 & 40  \\
            \hline
            %Contrastive \cite{hadsell2006dimensionality} &  - & - & - & - & - & - & - & - \\
            %Margin \cite{wu2017sampling} &  - & - & - & - & - & - & - & - \\
            NSM \cite{zhaiclassification2019} & 73.8 & 88.1 & 95.0 & - & - & - & - & - \\
            MS \cite{wang2019multi} &  78.2 & 90.5 & 96.0 & 98.7 & 89.7 & 97.9 & 98.5 & 99.1 \\
            SoftTriple \cite{qian2019softtriple} &  78.3 & 90.3 & 95.9 & - & - & - & - & -   \\
            HORDE$^\dagger$ \cite{JACOB_2019_ICCV} &  80.1 & 91.3 & 96.2 & 98.7 & 90.4 & 97.8 & 98.4 & 98.9 \\   
            %Proxy-Anchors \cite{kim2020proxy} &  79.1 & 90.8 & 96.2 & 98.7 & 91.5 & 98.1 & 98.8 & 99.1 \\
            DiVA \cite{milbich2020diva} &  78.1 & 90.6 & - & - & - & - & - & - \\
            \rowcolor[gray]{.9}
            Triplet & $75.8_{\pm0.1}$	& $87.9_{\pm0.1}$ & $94.1_{\pm0.1}$	& $97.6_{\pm0.1}$
            & $88.2_{\pm0.1}$	& $96.7_{\pm0.1}$ & $97.6_{\pm0.1}$	& $98.3_{\pm0.1}$ \\
            \rowcolor[gray]{.9}    
            Triplet+$L_2$Norm &  
            $79.1_{\pm0.1}$ & $90.9_{\pm0.1}$ & $96.3_{\pm0.1}$ & $98.8_{\pm0.1}$ &
            $90.1_{\pm0.1}$ & $97.8_{\pm0.1}$ & $98.6_{\pm0.0}$ & $99.1_{\pm0.1}$\\
            \rowcolor[gray]{.9}
            %Ours$^{512}$ & 68.4 & 78.2 & 85.6 & 91.2 & 87.7 & 92.6 & 95.3 & 97.3 \\
            Triplet+MDR &
            $\textbf{80.1}_{\pm0.0}$ & $\textbf{91.4}_{\pm0.1}$ & $\textbf{96.4}_{\pm0.1}$ & $\textbf{98.8}_{\pm0.1}$ &
            $\textbf{90.5}_{\pm0.1}$ & $\textbf{97.9}_{\pm0.1}$ & $\textbf{98.5}_{\pm0.1}$ & $\textbf{99.1}_{\pm0.1}$
            \\
            \rowcolor[gray]{.9}
            Triplet+MDR$^\dagger$&  
            $\textbf{80.8}_{\pm0.1}$ & $\textbf{91.9}_{\pm0.1}$ & $\textbf{96.7}_{\pm0.0}$ & $\textbf{98.9}_{\pm0.0}$ &
            $\textbf{91.3}_{\pm0.1}$ & $\textbf{98.2}_{\pm0.1}$ & $\textbf{98.8}_{\pm0.0}$ & $\textbf{99.3}_{\pm0.0}$ \\
           \hline
           \end{tabular}
		}
    	\caption{
    	Recall@\textit{K} comparison with state-of-the-art methods. 
        The baseline methods and MDR are grouped in the gray-colored rows.
        $\dagger$ indicates that the model is trained and tested with large images of $256\times256$ following the setting of \cite{JACOB_2019_ICCV}. 
        We round reported values to the first decimal place.
        %Triplet+$L_2$Norm \cite{schroff2015facenet}
    	}
		\label{tab:exp}
		\end{center}
\end{table*}
\endgroup

\noindent\textbf{Loss Function.} 
The proposed MDR can be applied to any loss functions $\mathcal{L}_{\text{DML}}$ such as Contrastive loss \cite{bromley1994signature}, Triplet loss \cite{schroff2015facenet} and Margin loss \cite{wu2017sampling}.
We mostly adopted Triplet loss as baseline for our experiments:
\begin{equation}
    \mathcal{L}_{\text{Triplet}}
    =
    \frac{1}{\left| {\mathcal{T}} \right| }
    \sum_{(e^a, e^p, e^n) \in \mathcal{T}} 
    \left[
    d(e^a, e^p)
    - d(e^a, e^n) + m \right] _ {+},
\end{equation}
where $\mathcal{T}$ is a set of triplets of an anchor $e^a$, a positive $e^p$, and a negative $e^n$ sampled from a mini-batch. $m$ is a margin.
The final loss function $\mathcal{L}$ is defined as the sum of $\mathcal{L}_{\text{DML}}$ and $\mathcal{L}_{\text{MDR}}$ with a multiplier $\lambda$ that balances the losses: 
\begin{equation}
    \mathcal{L} = \mathcal{L}_{\text{DML}} + \lambda \mathcal{L}_{\text{MDR}}.
\end{equation}
$\mathcal{L}_{\text{DML}}$ optimizes the model by minimizing the distance of positive pairs and maximizing the distance of negative pairs.
$\mathcal{L}_{\text{MDR}}$ regularize the pairwise distances by constraining the distances with multiple levels.
The embedding network is trained simultaneously with different objectives.
\smallbreak

\noindent\textbf{Embedding Normalization Trick for MDR.}
In our learning procedure, $L_2$ Normalization ($L_2$ Norm) is not adopted because it can disturb the proper regularization effect of MDR.
However, the lack of $L_2$ Norm can cause difficulty in finding appropriate hyper-parameters of $\mathcal{L}_\text{DML}$ such as margin $m$ in Triplet loss, because any prior knowledge of the scale of embedding vectors is not given.
To overcome the difficulty, we normalize $\mathcal{L}_{\text{DML}}$ by dividing the embedding vectors $e$ by $\mu$ during the training stage, such that the expected pairwise distance is one: $\mathop{\mathbb{E}}\big[d(\frac{e_i}{\mu},\frac{e_j}{\mu})\big]=1$.
We adopt this trick on several loss functions such as Constrastive loss \cite{hadsell2006dimensionality}, Margin loss \cite{wu2017sampling}, and Triplet loss in our experiments.

\section{Experiments}

% here is changed
To show the effectiveness of MDR and its behaviors, we extensively perform ablation studies and experiments.
We follow the standard evaluation protocol and data splits proposed in \cite{Song2016DeepML}.
For an unbiased evaluation, we conduct 5 independent runs for each experiment and report the mean and the standard deviation of them.
% All experiments are conducted on a single NVIDIA V100 with 32GB of memory.
\smallbreak

\noindent\textbf{Datasets.}
We employ the four standard datasets of deep metric learning for evaluations: CUB-200-2011 \cite{CUB-200} (CUB-200), Cars-196 \cite{Cars-196}, Stanford Online Product \cite{Song2016DeepML} (SOP) and In-Shop Clothes Retrieval \cite{InShop} (In-Shop).
CUB-200 has 5,864 images of first 100 classes for training and 5,924 images of the rest classes for evaluation.
Cars-196 has 8,054 images of first 98 classes for training and 8,131 images of the rest classes for evaluation.
SOP has 59,551 images of 11,318 classes for training and 60,502 images of the rest classes for evaluation.
In-Shop has 25,882 images of 3,997 classes for training, and the remaining 7,970 classes with 26,830 images are partitioned into two subsets (query set and gallery set) for evaluation.

\subsection{Implementation Details}

\noindent\textbf{Embedding Network.}
All the compared methods and our method use the Inception architecture with Batch Normalization (IBN) \cite{ioffe2015batch} as a backbone network.
IBN is pre-trained for ImageNet ILSVRC 2012 dataset \cite{deng2009imagenet} and then fine-tuned on the target dataset.
We attach a fully-connected layer, where its output activation is used as an embedding vector, after the last pooling layer of IBN.
For models trained with MDR, $L_2$ Norm is not applied to the embedding vectors because it disturbs the effect of the regularization.
For a fair comparison with the conventional implementation of Triplet loss \cite{schroff2015facenet} that is used as a baseline, we apply $L_2$ Norm to those models.
\smallbreak

\noindent\textbf{Learning.} 
We employ Adam \cite{kingma2014adam} optimizer with a weight decay of $10^{-5}$.
For CUB-200 and Cars-196, a learning rate and the size of mini-batch are set to $5\cdot10^{-5}$ and 128.
For SOP and In-Shop, a learning rate and the size of mini-batch are set to $10^{-4}$ and 256.
We mainly apply our method to Triplet loss \cite{schroff2015facenet}.
As a triplet sampling method, we employ the distance weighted sampling \cite{wu2017sampling}.
The margin $m$ of Triplet loss is set to 0.2.
We summarized the hyper-parameters of MDR:
the configuration of the levels is initialized to three levels of $\{-3, 0, 3\}$, and the momentum $\gamma$ is set to $0.9$.
$\lambda$ is set differently for each dataset: $0.6$ for CUB-200, $0.2$ for Cars-196 and $0.1$ for SOP and In-Shop.
For most of the datasets, $\lambda$ of $0.1$ is enough to improve a given model; on CUB-200, a strong regularization is more effective because it is a small dataset with only 5,864 training images where a model may easily suffer from overfitting.
Those hyper-parameters are not very sensitive to tune, and we explain the effects of each hyper-parameter in the ablation studies at Section \ref{sec:abl}.

\smallbreak

\noindent\textbf{Image Setting.}
%delete zhai
During training, we follow the standard image augmentation process \cite{Song2016DeepML,wang2019multi} with the following order: resizing to $256\times256$, random cropping, random horizontal flipping, and resizing to $224\times224$.
For evaluation, images are center-cropped.

\subsection{Comparison with State-of-the-art Methods}

We show the comparison of MDR and the recent state-of-the-art methods (Table \ref{tab:exp}).
All compared methods use embedding vectors of 512 dimensionality.
Our baseline model is trained by Triplet loss without $L_2$ Norm (Triplet) and we also report the conventional Triplet with $L_2$ Norm (Triplet+$L_2$ Norm).
The lack of constraints of $L_2$ Norm on the embedding space results in poor generalization performance, and it is known that Triplet loss is effective when $L_2$ Norm is applied \cite{schroff2015facenet}.
However, the models with MDR outperform the Triplet+$L_2$ Norm models on all the datasets.
Those results prove the effectiveness of the proposed distance-based regularization.

\smallbreak

\noindent\textbf{Experimental Results.} MDR improves performance on all the datasets, and, in particular, the improvements are significantly high on the small-sized datasets.
For CUB-200, MDR improves 3.7 percentage points on Recall@1 compared to the conventional Triplet+$L_2$ Norm; 
the result is 11.5 percentage points higher than Recall@1 of the Triplet.
For Cars-196, MDR improves 8.7 percentage points on Recall@1 compared to the conventional Triplet+$L_2$ Norm; the result is 12.3 percentage points higher than Recall@1 of the Triplet. 
MDR also improves the recall performance compared to the baselines on SOP and In-Shop.
Moreover, our method significantly outperforms the other state-of-the-art methods in all recall criteria for all datasets.

\begin{figure}[t]
	\begin{center}
		\subfloat[Dimensionality]{
		    \label{fig:discuss_a}
			\includegraphics[width=6.5cm]{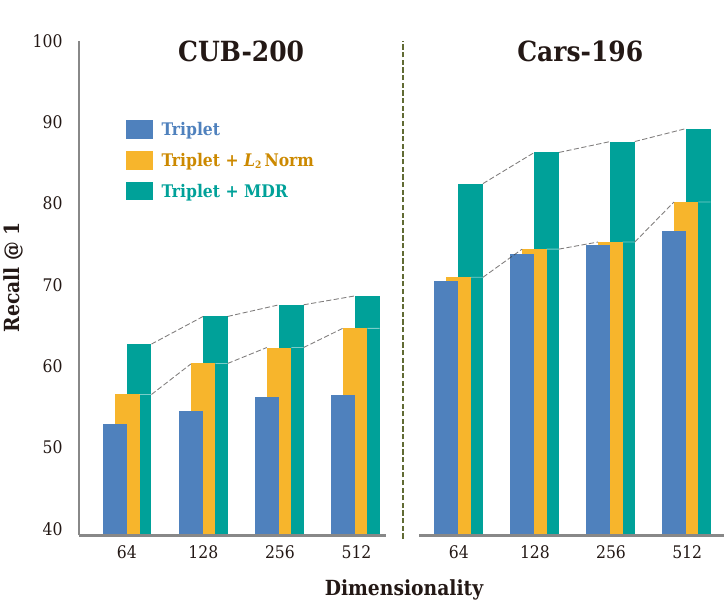}
		}	
		\\
		\subfloat[Learning Curves]{
		 \label{fig:discuss_b}
			\includegraphics[width=6.5cm]{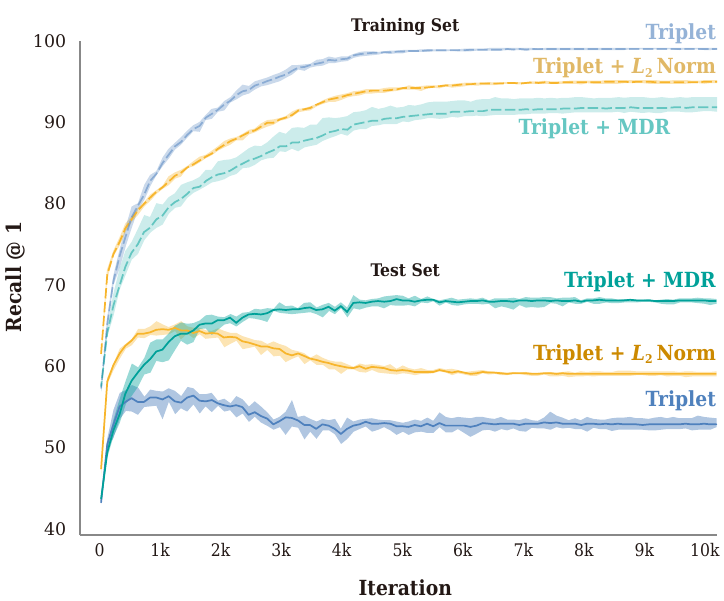}
		}
	\end{center}
	\caption{ 
    (a) compares the three methods on various dimensionalities of the embedding vector on CUB-200 and Cars-196.
    (b) shows the learning curves of the three methods for the training and test set on CUB-200.
  }
	\label{fig:discuss}
\end{figure}

\begingroup
\newcolumntype{C}[1]{>{\centering\let\newline\\\arraybackslash\hspace{0pt}}m{#1}}
\begin{table}[t!]
    \small
	\begin{center}
		\subfloat[Backbone Network\label{tab:abl_a} ]
		{
			\centering
			\def\arraystretch{1.2}
			\begin{tabular}{C{1.7cm}|cc|cc} 
           \hline
            & \multicolumn{2}{c|}{CUB-200} & \multicolumn{2}{c}{Cars-196} \\
            \cline{2-5}
            MDR & & $\checkmark$ &  & $\checkmark$ \\
            \hline
             {R18} & ${51.2}_{\pm0.5}$ & ${63.6}_{\pm0.5}$ & ${63.4}_{\pm1.0}$ & ${82.3}_{\pm0.2}$ \\
             {R50 } & ${58.5}_{\pm0.4}$ & ${65.8}_{\pm0.3}$ & ${77.5}_{\pm0.4}$ & ${87.6}_{\pm0.2}$  \\
             {IBN} & ${57.3}_{\pm0.7}$  & ${68.8}_{\pm0.5}$ & ${76.2}_{\pm0.6}$ & ${88.5}_{\pm0.3}$ \\
           \hline
           \end{tabular}
		} % R18 \cite{he2016deep} IBN \cite{ioffe2015batch}
		\\
		\subfloat[Loss Function\label{tab:abl_b} ]
		{
			\centering
			\def\arraystretch{1.2}
			\begin{tabular}{C{1.7cm}|cc|cc} 
           \hline
            & \multicolumn{2}{c|}{CUB-200} & \multicolumn{2}{c}{Cars-196} \\
            \cline{2-5}
            MDR & & $\checkmark$ &  & $\checkmark$ \\
            \hline
             {Contrastive} & ${63.9}_{\pm0.3}$ & ${65.6}_{\pm0.2}$ & ${83.2}_{\pm0.1}$ & ${86.1}_{\pm1.0}$ \\
            {Margin} & ${59.3}_{\pm0.5}$ & ${67.5}_{\pm0.3}$ & ${79.1}_{\pm0.3}$ & ${88.2}_{\pm0.4}$  \\
            {Triplet} & ${57.3}_{\pm0.7}$  & ${68.8}_{\pm0.5}$ & ${76.2}_{\pm0.6}$ & ${88.5}_{\pm0.3}$ \\
           \hline
           \end{tabular}
		} % Contrastive \cite{hadsell2006dimensionality}
		% Margin \cite{wu2017sampling}
		% Triplet \cite{schroff2015facenet}
		\\
		\subfloat[Level Configuration\label{tab:abl_c} ]
		{
			\centering
			\def\arraystretch{1.2}
			\begin{tabular}{C{3.2cm}|C{2cm}C{2cm}} 
            \hline
            & \multicolumn{2}{c}{CUB-200} \\
            \cline{2-3}
            & Fixed & Learnable \\
            \hline
            %$\{0\}$                 & ${67.9}_{\pm0.1}$ & ${68.0}_{\pm0.5}$ \\
            $\{-1, 0, 1\}$          & ${64.4}_{\pm0.4}$ & ${64.9}_{\pm0.4}$ \\
            $\{-2, 0, 2\}$          & ${67.9}_{\pm0.2}$ & ${68.2}_{\pm0.5}$ \\
            $\{-3, 0, 3\}$          & ${68.2}_{\pm0.1}$ & ${68.8}_{\pm0.5}$ \\
            $\{-3, -1, 0, 1, 3\}$   & ${64.0}_{\pm0.4}$ & ${64.9}_{\pm0.1}$ \\
            $\{-3, -2, 0, 2, 3\}$   & ${67.8}_{\pm0.4}$ & ${67.9}_{\pm0.3}$ \\
            $\{-4, -2, 0, 2, 4\}$   & ${67.8}_{\pm0.3}$ & ${67.8}_{\pm0.2}$ \\
            $\{-6, -3, 0, 3, 6\}$   & ${68.4}_{\pm0.1}$ & ${68.7}_{\pm0.5}$ \\
           \hline
           \end{tabular}
		}
    	\caption{
            % Recall@1 comparison with various backbone networks and loss functions on CUB-200 \cite{CUB-200} and Cars-196 \cite{Cars-196} datasets, and with level configurations on CUB-200 \cite{CUB-200}.
            % The models of (a) are trained with Triplet loss.
            % The models of (b) use IBN as the backbone network. 
            % In (a) and (b), a column with $\checkmark$ indicates that the models are trained with MDR.
            Recall@1 comparison with various backbone networks, loss functions, and level configurations.
            The models of (a) are trained with Triplet loss.
            The models of (b) use IBN as the backbone network. 
            In (a) and (b), a column with $\checkmark$ indicates that the models are trained with MDR.
    	}
		\end{center}
\end{table}
\endgroup

\begingroup
\newcolumntype{C}[1]{>{\centering\let\newline\\\arraybackslash\hspace{0pt}}m{#1}}
\begin{table}
	\begin{center}
			\centering
			\def\arraystretch{1.2}
        	\begin{tabular}{C{3.1cm}|C{2cm}C{2cm}} 
           \hline
            & \multicolumn{2}{c}{CUB-200} \\
            \cline{2-3}
            $L_2$ Norm at Inference & & $\checkmark$  \\
            \hline
            {Triplet} & ${57.3}_{\pm0.7}$ & ${51.5}_{\pm1.0}$  \\
            {Triplet+MDR} & ${68.8}_{\pm0.5}$  & ${68.2}_{\pm0.4}$ \\
           \hline
           \end{tabular}
    	\caption{
            % Recall@1 comparison with the effect of $L_2$ Norm in inference time on CUB-200 \cite{CUB-200} dataset. 
            % A column with $\checkmark$ indicates that the trained models are evaluated with $L_2$ Norm.
            Recall@1 comparison with the effect of $L_2$ Norm at inference time for the models trained without $L_2$ Norm.
            A column with $\checkmark$ indicates that the trained models are evaluated with $L_2$ Norm.
    	}
		\label{tab:norm}
		\end{center}
\end{table}

\begin{figure}[t]
	\begin{center}
		\subfloat[ {Expectation of Two-Norm: $E\left[\norm{e}_2\right]$} \label{fig:norm_a} ]
		{
			\includegraphics[width=7cm]{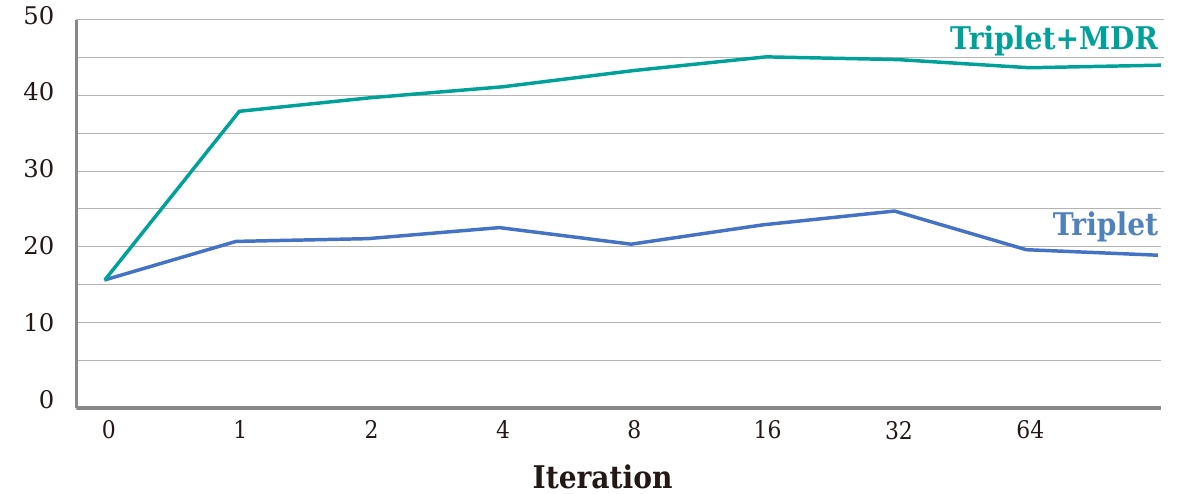}
		} 
		\\
		\subfloat[Coefficient of Variation of Two-Norm: {$\frac{Var\left[\norm{e}_2\right]}{E\left[\norm{e}_2\right]}$} \label{fig:norm_b} ]
		{
			\includegraphics[width=7cm]{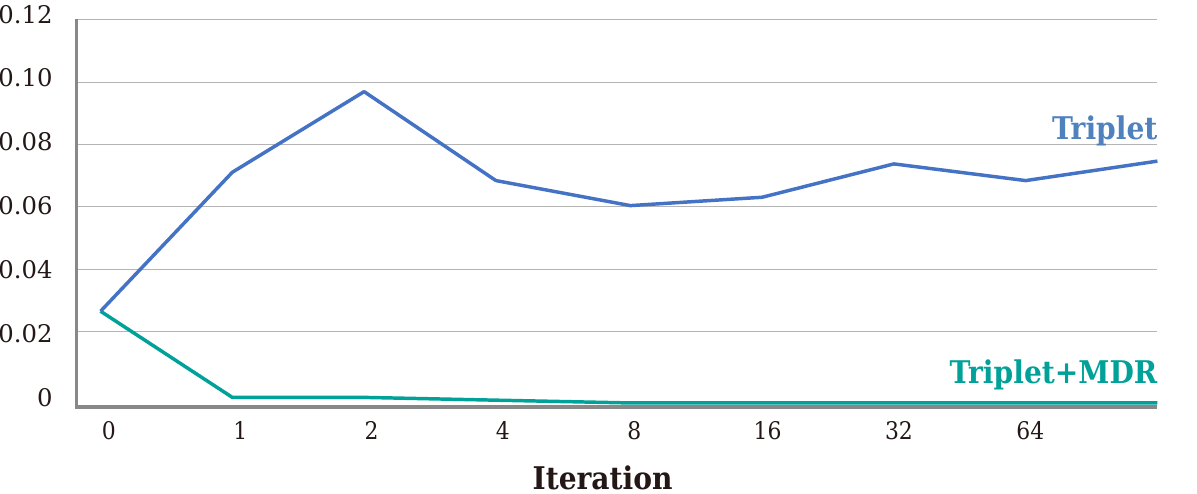}
		} 
	\end{center}
	\caption{ 
    % (a) compares the expectation of the scales of the embedding vectors for the test set on CUB-200 \cite{CUB-200}.
    % (b) compares the normalized variation of the scales of the embedding vectors for the test set on CUB-200 \cite{CUB-200}.
    %[자리]Analysis of the two-norm of the embedding vectors from the models trained without $L_2$ Norm.
    (a) compares the expectation of the two-norm of the embedding vectors for the test set on CUB-200.
    (b) compares the coefficient of variation of the two-norm of the embedding vectors for the test set on CUB-200.
    }
\end{figure}
\endgroup

\subsection{Ablation Studies}
\label{sec:abl}
We extensively perform ablation studies on the behaviors of the proposed MDR.
\smallbreak

\noindent\textbf{Backbone Network.} 
MDR can be widely applicable to any backbone networks (Table \ref{tab:abl_a}). 
We apply MDR on IBN \cite{ioffe2015batch}, ResNet18 (R18) and ResNet50 \cite{he2016deep} (R50), and achieve significant improvements for all backbone networks.
Especially, a light-weight backbone, R18, with MDR even outperforms the baseline models with a heavy-weight backbone such as R50 and IBN on both datasets.

\smallbreak

\noindent\textbf{Loss Function.}
Our MDR also can be widely applicable to any distance-based loss function (Table \ref{tab:abl_b}).
We apply MDR on Constrastive loss \cite{hadsell2006dimensionality}, Margin loss \cite{wu2017sampling} and Triplet loss.
MDR achieves significant improvements for all loss functions.
\smallbreak

\begin{figure*}[t]
	\begin{center}
		\subfloat[Triplet]{
			\label{fig:visualization_a}
			\includegraphics[height=4.5cm]{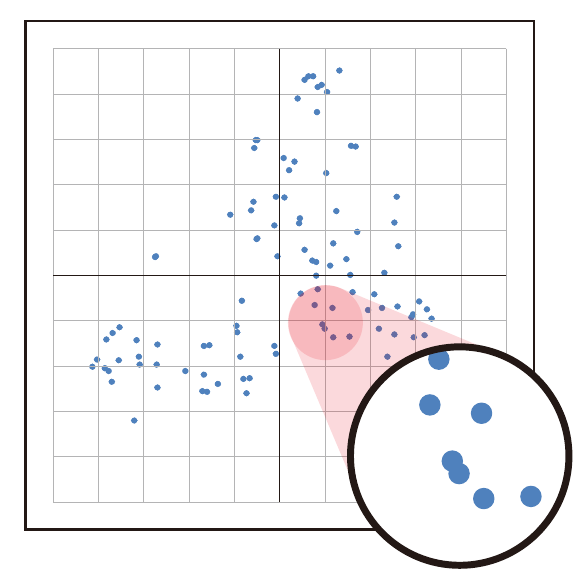}
		}
		\subfloat[Triplet + $L_2$ Norm]{
			\label{fig:visualization_b}
			\includegraphics[height=4.5cm]{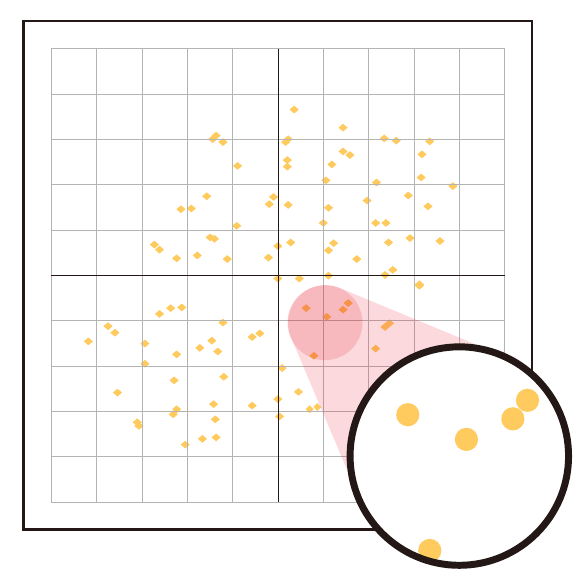}
		}	
		\subfloat[Triplet + MDR	\label{fig:visualization_c}]{		
			\includegraphics[height=4.5cm]{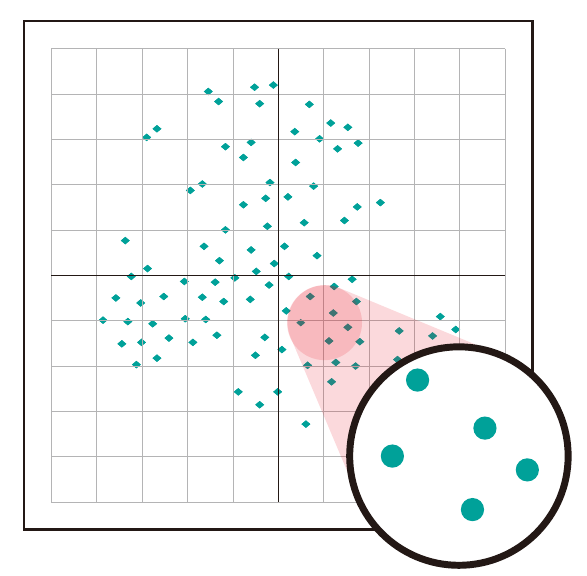}
		}	
	\end{center}
	\caption{
	Class centers in the embedding space of two models trained without MDR (Triplet \& Triplet+$L_2$ Norm) and one model trained with MDR (Triplet+MDR).
	We visualize using t-SNE \cite{t_SNE} on CUB-200.
	%We visualize using t-SNE \cite{t_SNE} on the test set of CUB-200 \cite{CUB-200}.
  }
	\label{fig:visualization}
\end{figure*}

\begin{figure*}[t]
	\begin{center}
		\includegraphics[width=16cm]{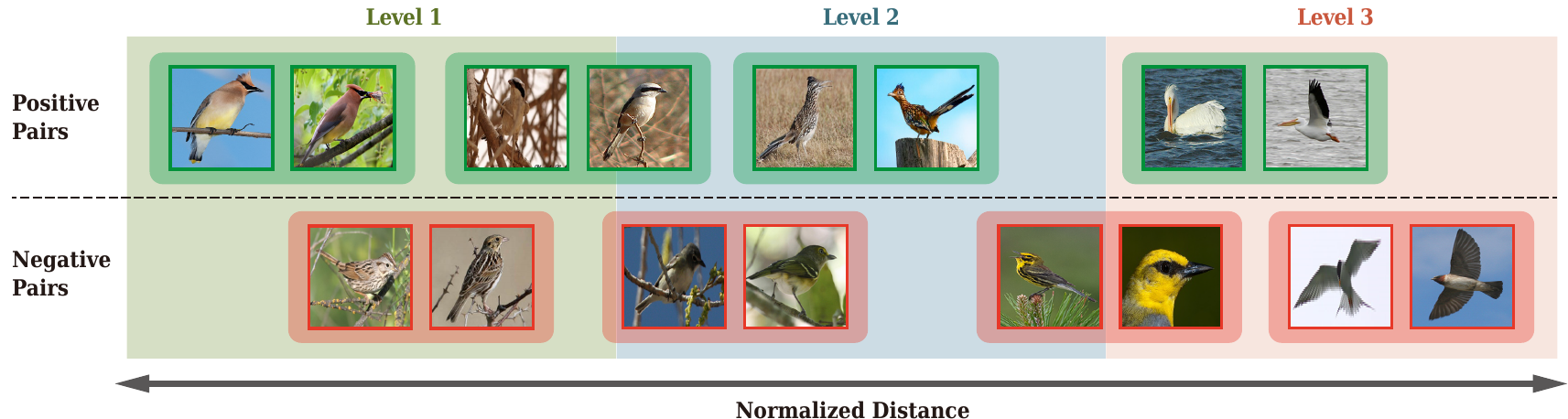}
	\end{center}
	\caption{
        Visualization of assigned positive and negative pairs at each level on CUB-200. 
        Regardless of positive or negative pair, a visually close pair is assigned to level 1, and a visually distant pair is assigned to level 3; even the same birds of the same species can be varying in appearance by the difference in perspectives, poses, and environments.
  }
	\label{fig:qa}
\end{figure*}

\noindent\textbf{Level Configuration $\mathcal{S}$.}
Even though the levels are learnable, we should properly set the number of levels and the initial values of levels.
We perform experiments on various initial configurations of levels and validate the importance of the learnability of levels (Table \ref{tab:abl_c}).
From the experiments, we find that a sufficiently spaced configuration is better than a tightly spaced configuration; $\{-3,0,3\}$ is better than $\{-1,0,1\}$, and a configuration of three levels is sufficient.

\subsection{Discussion}

\noindent\textbf{Effectiveness in Small Dimensionality.}
We perform an experiment on various dimensionalities of embedding vector such as $64$, $128$, $256$, and $512$.
MDR significantly improves the Recall@1 of the models, especially in small dimensionality.
In the experiment, our MDR only with 64 dimensionality is similar to or surpasses the performance of other methods with 512 dimensionality (Figure \ref{fig:discuss_a}).
The result indicates that our MDR constructs a highly efficient embedding space in compact dimensionality.
Moreover, the improvements are larger compared to Triplet+$L_2$ Norm for all dimensionality.
\smallbreak

\noindent\textbf{Prevention of Overfitting as Regularizer.}
We investigate the learning curves of three models: Triplet, Triplet+$L_2$ Norm and Triplet+MDR (Figure \ref{fig:discuss_b}).
There are two crucial observations:
(1) on the training set, Triplet+MDR is less overfitted than the other two methods, but it shows the most high performance on the test set., 
(2) the recall of Triplet+MDR does not drop until the end of learning, unlike the other methods, which suffer from severe overfitting. 
These observations indicate that our MDR is an effective regularizer for DML.
\smallbreak

\noindent\textbf{Equalizing the Two-Norm of Embedding Vectors.}
We find that the embedding vectors of a model trained with MDR have almost the same two-norm (Figure \ref{fig:norm_a} and \ref{fig:norm_b}).
This shows that the embedding vectors are almost located on a hypersphere, even though the model is trained without $L_2$ Norm.
Therefore, the model trained with MDR achieves similar performance even if $L_2$ Norm is applied at inference time (Table \ref{tab:norm}).
This observation implies that MDR has similar effects of $L_2$ Norm at the end of the training, even though MDR is a distance-based regularization and $L_2$ Norm is norm-based regularization.
\smallbreak

\noindent\textbf{Discriminative Representation.} To show the effectiveness of our method, we visualize how MDR constructs an embedding space.
In the embedding space of Triplet and Triplet+$L_2$ Norm, the class centers are often aligned closely to each other (Figure \ref{fig:visualization_a} and \ref{fig:visualization_b}).
However, in an embedding space of Triplet+MDR, the class centers are evenly spaced with a large margin (Figure \ref{fig:visualization_c}).
This result indicates that MDR constructs a more discriminative representation than the conventional methods.
\smallbreak

\noindent\textbf{Qualitative Analysis on Level Assignment.}
In the step of the level assignment, a lower level indicates that the pairs are closely aligned in the embedding space and vice versa.
Most of the positive pairs are belonging to between level 1 and 2, and most of the negative pairs are belonging to between level 2 and 3.
However, hard-positive pairs may belong to level 3 while hard-negative also may belong to level 1 (Figure \ref{fig:qa}).
Therefore, levels are assigned to each pair regardless of given binary supervision.
The learning procedure tried to overcome the disturbance that pulls the distances to belonging levels by considering the various degrees of distances; this multi-level disturbance leads to the improvement of the generalization ability.
\smallbreak

\section{Conclusion}
We introduce a new distance-based regularization method that elaborately adjusts the pairwise distance into multiple levels for better generalization.
We prove the effectiveness of MDR by showing the improvements that greatly exceed the existing methods, and by extensively performing the ablation studies of its behaviors.
By applying our MDR, many methods can be significantly improved without any extra burdens at inference time.

\section*{Acknowledgement}
We would like to thank AI R\&D team of Kakao Enterprise for the helpful discussion.
In particular, we would like to thank Yunmo Park who designed the visual materials.
\\

\noindent\textbf{Potential Ethical Impact}
\\
Due to the gap between a training dataset and real-world data, 
it is important to build a reliable model with better generalization ability across the unseen dataset, \eg test set, for its practicality.
Our MDR is a regularization method to improve the generalization ability of a deep neural network on the task of deep metric learning.
As positive aspects, our method can be applied to many practical applications such as image retrieval and item recommendation. These applications are utilized for our conveniences and the proposed MDR can improve their performance more reliably.
We believe that our method does not have particular negative aspects because it is a fundamental method that assists conventional approaches to improve reliability on unseen datasets.

\small
\bibliography{egbib}

\end{document}